# Medical visual question answering using joint self-supervised learning

Yuan Zhou[1], Jing Mei[1], Yiqin Yu[1] and Tanveer Syeda-Mahmood[2]

[1] IBM Research China, Beijing, China
{zhxyuan, meijing, yuyiqin}@cn.ibm.com
[2] IBM Research, Almaden Research Center, San Jose, CA, USA
stf@us.ibm.com

**Abstract.** Visual Question Answering (VQA) becomes one of the most active research problems in the medical imaging domain. A well-known VQA challenge is the intrinsic diversity between the image and text modalities, and in the medical VQA task, there is another critical problem relying on the limited size of labelled image-question-answer data. In this study we propose an encoder-decoder framework that leverages the image-text joint representation learned from large-scaled medical image-caption data and adapted to the small-sized medical VQA task. The encoder embeds across the image-text dual modalities with self-attention mechanism and is independently pre-trained on the large-scaled medical image-caption dataset by multiple self-supervised learning tasks. Then the decoder is connected to the top of the encoder and fine-tuned using the small-sized medical VQA dataset. The experiment results present that our proposed method achieves better performance comparing with the baseline and SOTA methods.

**Keywords:** Visual Question Answering, Medical Imaging, Self-supervised Learning.

## 1 Introduction

Medical images are generally hard to be understood by most patients and their families, who naturally would ask questions about their health conditions, the detailed disease status and diagnosis evidence etc., besides the final diagnostic conclusion. Therefore, an AI-enabled system is expected to have the ability to answer questions about the specific medical images. For general purpose, the visual question answering (VQA) task has been proposed and implemented for answering the natural language questions according to the content of the reference image [1-3]. Imaginably, VQA technologies for medical imaging could empower the patients and their families, as well as providing a second opinion for clinicians.

In the medical VQA, training the deep neural networks from scratch has been constrained by the lack of labeled image-question-answer data. Fine-tuning the pre-trained model based on the large scale labeled datasets such as ImageNet [4] and MS-COCO [5] is also difficult due to the large domain shift between general images and medical images [6,7]. Fortunately, a large scale of medical image data has been generated from the radiological examinations in the clinical conditions. This kind of medical image data generally includes two components: medical image and medical reports [8-10]. The medical image provides the visual information and the medical report provides the textual information which captions the medical image in detail. This type of image-caption data motivates our approach to overcome the data limitation in medical VQA. We propose an encoder-decoder framework for the medical VQA task. The large-scaled medical image-caption data from the medical image reports can be used to pre-train an image-text encoder that learns an image-text joint embedding by self-supervised learning. Then a text decoder is coupled with the encoder to generate the answer sequence. The whole encoder-decoder framework can be fine-tuned with the small-sized medical image-question-answer data. The medical image-question data will be aligned closely in the latent space based on the pre-trained dual-modal encoder. The fine-tune process can be more sufficient and empowered to overcome the limitation of medical VQA data.

## 2    Related Work

The CLEFmedicine challenge proposed a medical VQA task in recent years, many different methods have been introduced in their review paper [11]. Most studies on medical VQA leverage the methods from the general VQA domain. Usually, the convolutional neural network (CNN) such as ResNet [12], Inception [13] or DenseNet [14] pre-trained on ImageNet is used as the image feature extractor. The question is fed into a recurrent neural network (RNN) for text embedding. Different feature fusion strategies such as MUTAN [15,16] with attention mechanisms such as stacked attention network (SAN) [17], multi-modal compact bilinear pooling (MCB) [18] and bilinear attention networks (BAN) [19] are introduced to learn the joint embedding. Training these models from scratch requires large-scaled medical VQA labeled data, while transferring the pre-trained model from general domain need to conquer the domain shift. Nguyen et al. introduce the meta-learning and auto-encoder to overcome the data limitation in the medical VQA task, but the method needs external data with handcrafted labels [20]. Differently, in this paper, we propose to leverage the image-text joint representation trained on large-scaled medical image-caption data and adapted to the medical VQA task. Our proposal of a self-supervised learning mechanism mitigates the demands of labeled data and reduces the domain shift than transferring models from the general domain.

## 3    Methods

### 3.1    Model Overview

As shown in Fig. 1(a), we leverage an encoder-decoder framework, accepting a pair of image and question as the input, with output of an answer. For a VQA task, the image and question are embedded into joint representation by the encoder and the decoder generates the answer sequence from this joint embedding. In detail, the proposed VQA model consists of three modules.

**Module 1: Single modal encoder.** In this module, an *image embedder* and a *text embedder* extract the respective embedding of the input image and text. Specifically, we use *Inception-V3* [21] as the backbone of the image embedder to extract the visual features. The feature maps of the last pooling layer are passed through a layer normalization (LN) to generate the image embedding. For the *text embedder*, texts are tokenized and each token is embedded as a word vector. We get the patch-level image features $V$ and the token-level text embedding $T$ as the output.

**Module 2: Image-text joint encoder.** This module is for getting the joint embedding from the individual embedded image and text from **Module 1**. Instead of concatenating the dual-modal embedding directly, we use a multi-layer self-attention Transformer encoder to learn a cross-modal joint embedding between the image and text embedding. As shown in Fig. 1(a), the image feature patches and text token embeddings are reformed to a sequence $S_{V+T}$ and fed to the Transformer encoder. The output embedding $E_{V+T}$ can jointly represent the image-text dual-modal information due to the self-attention mechanism.

**Module 3: Answer decoder.** The answer is generated by a sequential generating model (Transformer decoder) with accepting the joint embedding of the image-question pair. Instead of using a discriminative model (e.g. a classifier) to match the most probable answer from a closed-ended domain of all the answer candidates, we leverage a generative model to better address the open-ended questions which are more frequently observed in the medical cases. The attention information between input dual-modal embedding (V&Q) and the generated answer tokens (A) can overcome the challenge of long-range dependencies when generating the answer sequence.

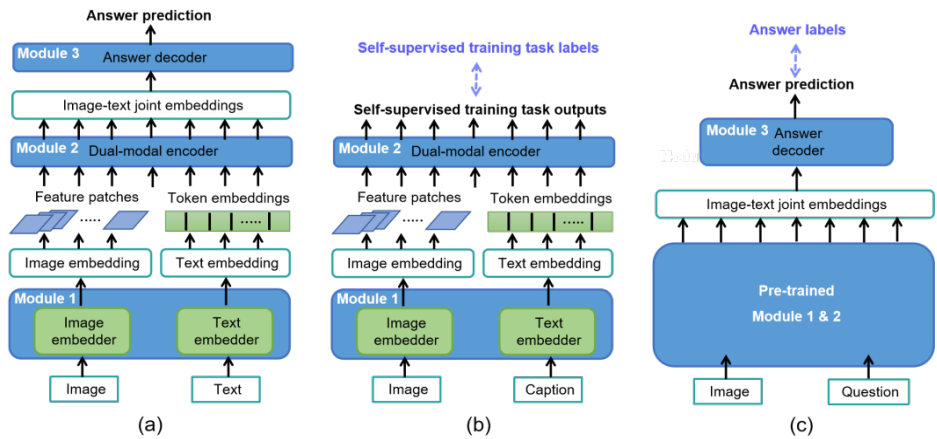

**Fig. 1.** The proposed medical VQA framework. (a) Model architecture and data flow. (b) Pre-training phase. Image-caption data is given as the input for self-supervised learning. (c) Fine-tuning phase. Image-question-answering data is given as the input for supervised learning.

### 3.2 Model Training

Training the encoder-decoder model is divided into two phases: pre-training and fine-tuning. Figs. 1(b) and (c) show the details of the two phases, separately. In the pre-training phase, we decouple the encoder (**Modules 1 and 2**) from the whole framework and perform a self-supervised training strategy using the large-scaled image-caption data. In the fine-tuning phase, the decoder (**Module 3**) is connected to the pre-trained encoder and then the whole model is trained on the small-sized image-question-answer data. The encoder accepts the image-text data, so it can be directly transferred from the image-caption data to the image-question data when the training phase is switched into fine-tuning from pre-training.

**Pre-training phase.** The pre-training phase aims at learning a joint embedding across the image and text modalities from the large-scaled image-caption data. The caption describes the detail of the corresponding image, including the information of imaging modality, organ location and abnormal regions. Therefore, the encoder can mine the alignments between the image and its caption in the latent space by training on this kind of data. Since the image-caption data does not have explicit labels, we design several self-supervised training tasks to guide the model to learn the joint contextualized embedding for medical images and captions.

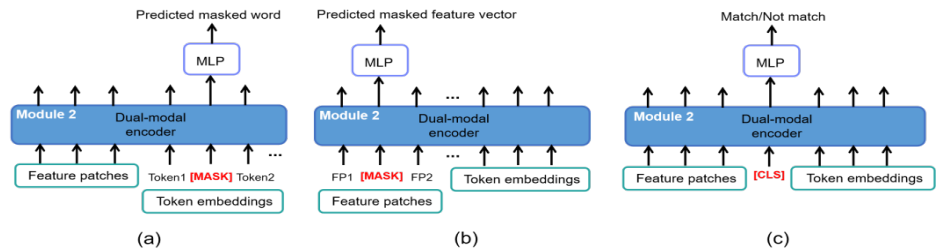

**Fig. 2.** Self-supervised training tasks in the pre-training phase. (a) Masked Word Prediction (MWP). (b) Masked Feature Regression (MFR). (c) Image Text Matching (ITM).

Three self-supervised training tasks are designed in the pre-training phase: Masked Word Prediction, Masked Feature Regression, and Image Text Matching, shown in Figs. 2(a), (b) and (c), respectively.

*Masked Word Prediction (MWP).* In this task, the feature patches $V$ and text token embeddings $T$ are reformed to a sequence made of feature patches $V = \{v_1, v_2, ..., v_M\}$ and token embeddings $T = \{t_1, t_2, ..., t_N\}$, each feature patch is a 1-D vector and represents the embedding of a region on the original image. We randomly mask the input words with a probability of 15% and replace the masked word token embeddings with a special token *[MASK]* (denoting the indices as $k \in N^K$). We input the feature patches and the word token embedding sequence with masked token embedding inside. The image-text joint encoder needs to predict the masked word token embeddings $T_k$ based on the observation of surrounding word token embeddings $T_{\backslash k}$ and feature patches $V$. To enhance the generalization of the joint encoder, we use a two-layer multi-layer perceptron (MLP) on top of the Transformer encoder to predict the results. The training loss is defined as the negative log-likelihood:

$$L_{MWP}(\theta_{en}) = -E_{(V,T) \sim D_{train\_ic}} \log P_{\theta_{en}}(T_k \mid T_{\setminus k}, V) \quad (1)$$

where $\theta_{en}$ represents the trainable parameters of encoder, each image-caption pair is sampled from the image-caption training dataset $D_{train\_ic}$.

*Masked Feature Regression (MFR).* Similar to the MWP task, we randomly mask the input feature patches $V$ with the probability of 15% and replace the masked feature patches $V_k$ by zero. In this task, the goal is to predict the masked feature patch given the surrounding feature patches $V_k$ and the word token embeddings $T$. For each single masked feature patch $V_k^i$, we also use a two-layer MLP to convert the corresponding Transformer output to a predicted feature vector $h(V_k^i)$. We apply an L2 regression loss between the predicted and the ground truth of the masked feature patch in this training task:

$$L_{MFR}(\theta_{en}) = -E_{(V,T) \sim D_{train\_ic}} f_{\theta_{en}}(V_k \mid V_{\setminus k}, T) \qquad where \quad f_{\theta_{en}}(V_k \mid V_{\setminus k}, T) = \sum_{i=1}^{K} \left\| h_{\theta_{en}}\left(V_k^i\right) - V_k^i \right\|_2^2 \quad (2)$$

*Image Text Matching (ITM).* In this task, an additional special token *[CLS]* is assigned at the breakpoint of the input dual modalities (Fig. 2(c)). A classification MLP is also applied on top of the Transformer, to classify if the input image and text matches with each other. If the input text is the caption of the input medical image, the ground truth of the output should be true. We denote the label as $y \in \{0,1\}$, indicating if the image-text pair is well matched or not. We replace the image or text in the original pair by the randomly-selected ones in the rest of the pairs to build the negative samples. The binary cross-entropy loss is used in this task:

$$L_{ITM}(\theta_{en}) = -E_{(V,T) \sim D_{train\_ic}} \left[ y \log f_{\theta_{en}}(V,T) + (1-y) \log\left(1 - f_{\theta_{en}}(V,T)\right) \right] \quad (3)$$

where $f_{\theta_{en}}(V,T)$ is the classification result from the MLP for a given image-caption pair $(V,T)$.

**Fine-tuning phase.** The labeled image-question-answer data is used to fine-tune the whole model for answer generating. The Transformer decoder is connected to the top of the pre-trained encoder. From the bottom, the image-question pair is fed into the encoder and the decoder generates the token sequence step-by-step. We apply the teacher forcing at each step $t$, the corresponding label token $y_t$ is fed to the decoder together with the image-question joint embedding. The whole model is fine-tuned by minimizing the negative log-likelihood of the generated tokens:

$$L_{VQA}(\theta) = -E_{(V,Q,A) \sim D_{train\_vqa}} \sum_{t=1}^{S} \log P_{\theta}(y_t \mid y_1, y_2, ..., y_{t-1}) \quad (4)$$

where $\theta$ is the trainable parameters of the whole model, each image-question-answer triple is sampled from the training dataset $D_{train\_vqa}$.

In the fine-tuning phase, a special token *[EOS]* (end-of-sentence) is added to the end of the labeled answer sentences as the endpoint. The decoder will stop generating the answer tokens once the *[EOS]* token is generated. Based on the pre-trained encoder, the whole model is fine-tuned with the VQA-MED dataset as described in section 4.

## 4    Experiments

### 4.1    Datasets and Implementation Details

**Datasets.** All the datasets are from the ImageCLEFmedical public challenges in ImageCLEF 2019 [11]. The image-caption dataset used for pre-training is from the image caption task and the image-question-answer dataset used for fine-tuning is from the VQA-MED-2019 task. **Pre-training dataset:** The medical image-caption dataset includes 70786 image-caption pairs. Each image matches one unique caption. The images cover most of the mainstream imaging modalities. We divide the dataset into 56629 pairs for training and 14157 pairs for test/validation. **Fine-tuning dataset:** The medical image-question-answering dataset (VQA-MED dataset) includes 3700 images and 14792 question-answer (QA) pairs for training and 500 images and 2000 QA pairs for test/validation. Each image is assigned with more than one question.

**Model details.** All the input images are resized to the shape of 299×299 and fed into the *Inception-V3* pre-trained on the ImageNet data. The feature maps from the output of the last layer (8×8×2048) are reshaped to 64×2048. Each text is trimmed into a 40-word sequence by padding zero to which the sequence length is less than 40. Each word in the text is represented by a 300-D word embedding by GloVe [24]. The image feature maps and word embedding are both passed to an extra 128-D embedding layer, each single feature patch and word token is converted to a 128-D vector. Finally the image is represented by a sequence of 64 feature patches and the text is embedded in a 40-token sequence. The text sequence is attached to the end of the image feature patch sequence when they are fed into the joint encoder. The joint encoder and the answer decoder are both an 8-layer Transformer with 8 self-attention heads. The whole model is implemented based on Tensorflow (v2.0.0) [25].

**Training details.** All the training and testing experiments are conducted on Nvidia Tesla P100 GPUs (16GB VRAM; PCIe connection) with distributed computing. In the pre-training phase, the three self-supervised training tasks (MWP, MFR and ITM) are conducted in parallel to consolidate the training performance. We pre-train the encoder using the batch size of 32 over 100K steps. We minimize the summation of the 3 training losses in Eqs. (1), (2) and (3) using the Adam optimizer with a learning rate of 1e-4 and weight decay of 0.001. In the fine-tuning phase, we use image-question-answer triples for training the whole model (encoder and decoder) with a batch size of 32. The Adam optimizer with a learning rate of 3e-4 and weight decay of 0.001 is used over 30K iterations.

**Evaluation details.** We follow the evaluation methods in the VQA task of ImageCLEFmedical 2019 Challenge [11]. Before running the evaluation metrics, a rule-based pre-processing is applied to the generated answer from the model. First, each character in the answer is converted to lower-case. Second, all punctuations are removed and the answer is tokenized to individual words. VQA accuracy and BLEU [26] are selected as the evaluation metrics. The VQA accuracy quantifies the fraction of the predicted answer that exactly matches the ground truth.

## 4.2 Ablation study

We evaluate the effects of the pre-training phase and the image-text joint encoder. In the first experiment, we compare the VQA results of the models whether pre-trained on the image-caption dataset or not. In the second experiment, we present the results of models with or without the image-text joint encoder, to evaluate whether the encoder can learn across the dual modalities and contribute to the answer generating process. As a control group, we remove the image-text joint encoder and input the individual embeddings of image and text to the answer decoder. The self-supervised training strategy to pre-train on the image-caption dataset cannot be used in the control group, we only use VQA dataset for training from scratch.

Table 1 shows the VQA accuracy and BLEU of the ablation study on the test set. "With/without joint encoder" means if the image-text joint encoder module has been used or not and "fine-tuning/train from scratch" means the self-supervised pre-training has been conducted or not. The first two rows show the results of two different setups in the first experiment. The pre-training on the image-caption dataset brings a large improvement in the VQA performance. This result validates the feasibility of the self-supervised pre-training phase that can leverage more information to overcome the data limitation in the medical VQA task. The second and third rows give the results in the second experiment. To align with the control group (the third row in Table 1), the whole framework with the joint encoder is also trained from scratch on the VQA dataset (the second row Table 1). The VQA performance has been improved with the introducing of the joint encoder module, even without the pre-training phase. From all the results in Table 1, the self-supervised pre-training provides the mechanism to learn and leverage the information from the large-scaled image-caption dataset, while the joint encoder module supports the learning across the image and text modalities.

## 4.3 Comparison with the benchmarks

We compare the proposed method with the benchmark methods. We select the top-2 methods on the ImageCLEFmedical challenge 2019 leader board [11] and several general VQA frameworks reported in [27], including SAN [17], MCB [18] and BAN [19] for comparing. The state-of-the-art (SOTA) is a meta-learning based method using bilinear attention networks (MAML+BAN) [20].

Table 2 presents the results of all the methods for comparison. Our proposed method with pre-training outperforms other benchmarks on both VQA accuracy and BLEU from the reported results. All the methods only use the VQA dataset for training except for our proposed method and MAML+BAN. The proposed method shows the worst results when it is only trained from scratch on the VQA dataset. The reason is that the proposed method has large-scaled trainable parameters that cause instability when training with such small-sized data. This phenomenon also presents that the embedding from self-supervised pre-training can empower the downstream VQA task. The SOTA method also uses external datasets except for the small-sized VQA dataset in the meta-learning phase. The difference is they need the dataset with handcrafted labels which are not necessary for our proposed method.

**Table 1.** VQA results on the test set in the ablation study.

| Model | VQA accuracy | BLEU |
|---|---|---|
| With joint encoder (pre-train+fine-tune) | 0.675 | 0.688 |
| With joint encoder (from scratch) | 0.432 | 0.452 |
| Without joint encoder (from scratch) | 0.407 | 0.415 |

**Table 2.** VQA results of different methods on the test set.

| Method | VQA accuracy | BLEU |
|---|---|---|
| SAN [17] | 0.490 | Not reported |
| MCB [18] | 0.518 | Not reported |
| BAN [19] | 0.568 | Not reported |
| MAML+BAN (SOTA) [20] | 0.673 | Not reported |
| ImageCLEF rank-1 [11] | 0.624 | 0.644 |
| ImageCLEF rank-2 [11] | 0.620 | 0.640 |
| Proposed method (from scratch) | 0.432 | 0.452 |
| **Proposed method (pre-train)** | **0.675** | **0.688** |

## 5 Conclusion

In this work, we propose an encoder-decoder framework that can leverage the image-text joint representation trained on large-scaled medical image-caption data to the data-limited medical VQA task. The encoder embeds across the image-text dual modalities with self-attention mechanism and is independently pre-trained on the large-scaled medical image-caption dataset by multiple self-supervised learning tasks. Then the decoder is connected to the top of the encoder and fine-tuned with the encoder together using medical VQA dataset. The experiment results demonstrate that this method achieves better performance comparing with the baseline and SOTA methods. The joint encoder supports the cross-modal learning and bridge the semantic gap between image and text. The self-supervised learning mitigates the requirements of labeled data and reduces the domain shift than transferring models from the general domain. The generated decoder model can address both close-ended and open-ended questions. This self-supervised learning based image-text joint encoder can be leveraged to other visual reasoning tasks such as image-text retrieval, low-resource image caption and visual entailment.